\documentclass[11pt]{article}

\usepackage[preprint]{acl}

\usepackage{multirow} 
\usepackage{graphicx}
\usepackage{algorithm}
\usepackage{algorithmic}

\usepackage{hyperref}

\usepackage{float}

\usepackage{amssymb}  

\usepackage{times}
\usepackage{latexsym}
\usepackage{tcolorbox}
\usepackage{textcomp}
\tcbuselibrary{breakable}

\usepackage[T1]{fontenc}

\usepackage[utf8]{inputenc}

\usepackage{microtype}

\usepackage{inconsolata}

\usepackage{graphicx}
\usepackage{amsmath}

\title{Structure-Grounded Knowledge Retrieval via Code Dependencies for Multi-Step Data Reasoning}

\author{
 \textbf{Xinyi Huang}
\\
 Simon Fraser University
}

\begin{document}
\maketitle

\makeatletter
\renewcommand\thefootnote{} 
\makeatother

\footnotetext{
\textsuperscript{$\dagger$} Work done during the authors' internships at Microsoft.
}

\makeatletter
\renewcommand\thefootnote{\arabic{footnote}} 
\makeatother

\begin{abstract}
Selecting the right knowledge is critical when using large language models (LLMs) to solve domain-specific data analysis tasks. However, most retrieval-augmented approaches rely primarily on lexical or embedding similarity, which is often a weak proxy for the task-critical knowledge needed for multi-step reasoning.
In many such tasks, the relevant knowledge is not merely textually related to the query, but is instead grounded in executable code and the dependency structure through which computations are carried out.
To address this mismatch, we propose \textsc{SGKR} (Structure-Grounded Knowledge Retrieval), a retrieval framework that organizes domain knowledge with a graph induced by function-call dependencies. Given a question, \textsc{SGKR} extracts semantic input and output tags, identifies dependency paths connecting them, and constructs a task-relevant subgraph. The associated knowledge and corresponding function implementations are then assembled as a structured context for LLM-based code generation.
Experiments on multi-step data analysis benchmarks show that \textsc{SGKR} consistently improves solution correctness over no-retrieval  and similarity-based retrieval baselines for both vanilla LLMs and coding agents.


\end{abstract}

\section{Introduction}

Large language models (LLMs) have recently shown strong capabilities in generating executable programs for complex reasoning and data analysis tasks~\cite{codex,huang-etal-2024-da,chan2024mle-bench,hu2024infiagent,ds1000}.

However, domain-specific data analysis tasks often involve numerical computations, hard-coded business rules, and structured data operations such as filtering, joining, aggregation, and conditional transformation~\cite{finqa, ConvFinQA}, which makes code generation substantially more challenging. In such scenarios, the correctness of generated programs heavily depends on whether the model can access and apply the appropriate domain knowledge. 
Similar challenges have been observed in related structured generation tasks such as text-to-SQL. For example, in the popular BIRD benchmark~\cite{bird}, prior work has shown that providing LLMs with domain knowledge or ``evidence'' about the database --- such as explanations of abbreviated medical terms in column names or clarifications of subtle calculation rules like ratio definitions --- can drastically boost SQL generation accuracy~\cite{arming}. These examples highlight a broader issue: beyond general reasoning and coding ability, LLMs often fail due to the lack of specific domain knowledge needed to produce the correct program.

The knowledge gap challenge becomes even more pronounced in broader, multi-step data analysis tasks, where models may need to discover latent computational rules, business logic, or function-level dependencies that are essential for solving the task.
To mitigate this issue, recent work commonly adopts retrieval-augmented generation (RAG)~\cite{codebert,graphcodebert,Rani2024AugmentingCS,gnnrag}, where relevant knowledge stored in knowledge graphs or free-text documentation is retrieved and supplied to the model as additional context.

Most existing retrieval approaches rely primarily on lexical matching~\cite{Robertson1994OkapiAT} or embedding similarity~\cite{karpukhin2020dense,xiongapproximate} between the input query and a given knowledge base. More recent graph-based RAG methods organize retrieved information as graphs, but still require embedding similarity to identify relevant nodes or subgraphs~\cite{graphrag}. 
While effective for knowledge-intensive question answering, these approaches assume that relevant knowledge can be retrieved as independent text fragments. However, in data analysis tasks, solving a problem requires composing interdependent computational steps, where correctness depends not only on retrieving relevant knowledge, but also on retrieving the right set of functionally connected operations.


However, in many domain-specific data analysis tasks, the knowledge required to solve a problem is encapsulated implicitly in code. Furthermore, function-call dependencies provide a structural signal of knowledge relevance: if a function A invokes another function B, the dependency often reflects that the callee (B) contributes logic flow or knowledge required by the caller (A). Therefore, relevant knowledge may be connected through code dependencies, even when the functions may only be weakly related to the input query in lexical or embedding space.



Motivated by the observation above, we propose Structure-Grounded Knowledge Retrieval (\textsc{SGKR}), a framework that organizes domain knowledge as a code dependency graph derived from function-call relations. 
Given a new question, the system extracts key input and output entities from the query as semantic I/O tags --- such as data entities, attributes, and target derived quantities --- maps these tags to relevant functions input/output in the code graph, and identifies dependency paths among these functions to construct a task-relevant subgraph.

Our contributions are summarized as follows:

\begin{itemize}

\item We introduce \textbf{Structure-Grounded Knowledge Retrieval (\textsc{SGKR})}, a retrieval framework that organizes domain knowledge along with a code dependency graph and retrieves relevant knowledge through dependency paths between input and output entities.

\item To organize knowledge as a graph, we propose to structure domain knowledge based on its usage in code by associating knowledge with corresponding functions and organizing them through code dependency relations.


\item Experiments on multiple multi-step data analysis benchmarks demonstrate that \textsc{SGKR} significantly improves solution correctness over no-retrieval and similarity-based retrieval baselines while retrieving substantially less context.
\end{itemize}

\begin{figure*}[h]
  \includegraphics[width=\textwidth]{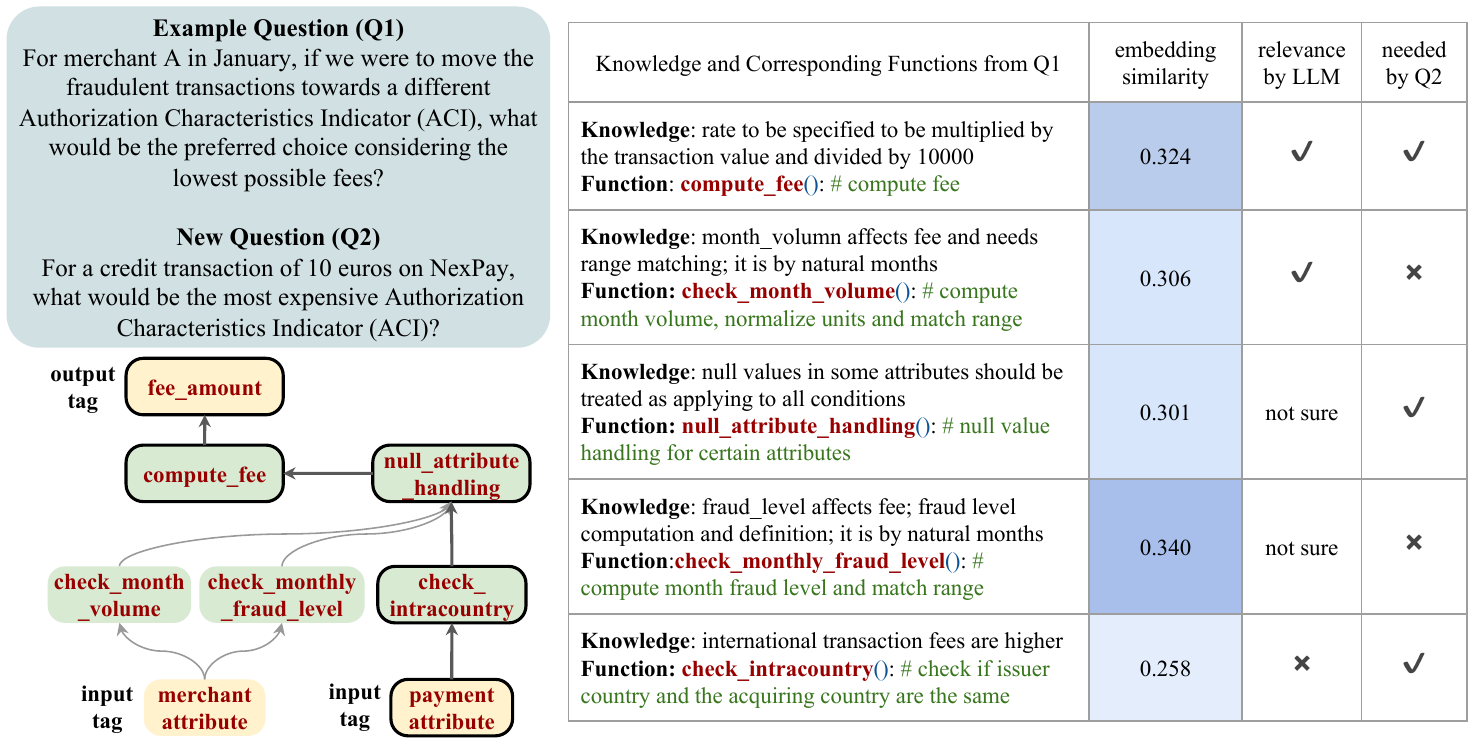}
  \caption{Comparison among (1) embedding-based similarity between the new task and candidate code snippets, (2) LLM (GPT-5.2) predictions of whether code snippets are required, and (3) ground-truth annotations indicating whether code snippets are actually needed. For similarity computation, we use the openai/text-embedding-3-large model, embedding both function definitions and their natural-language summaries.}
  \label{fig:motivation}
\end{figure*}

\begin{figure*}[t]
  \includegraphics[width=\textwidth]{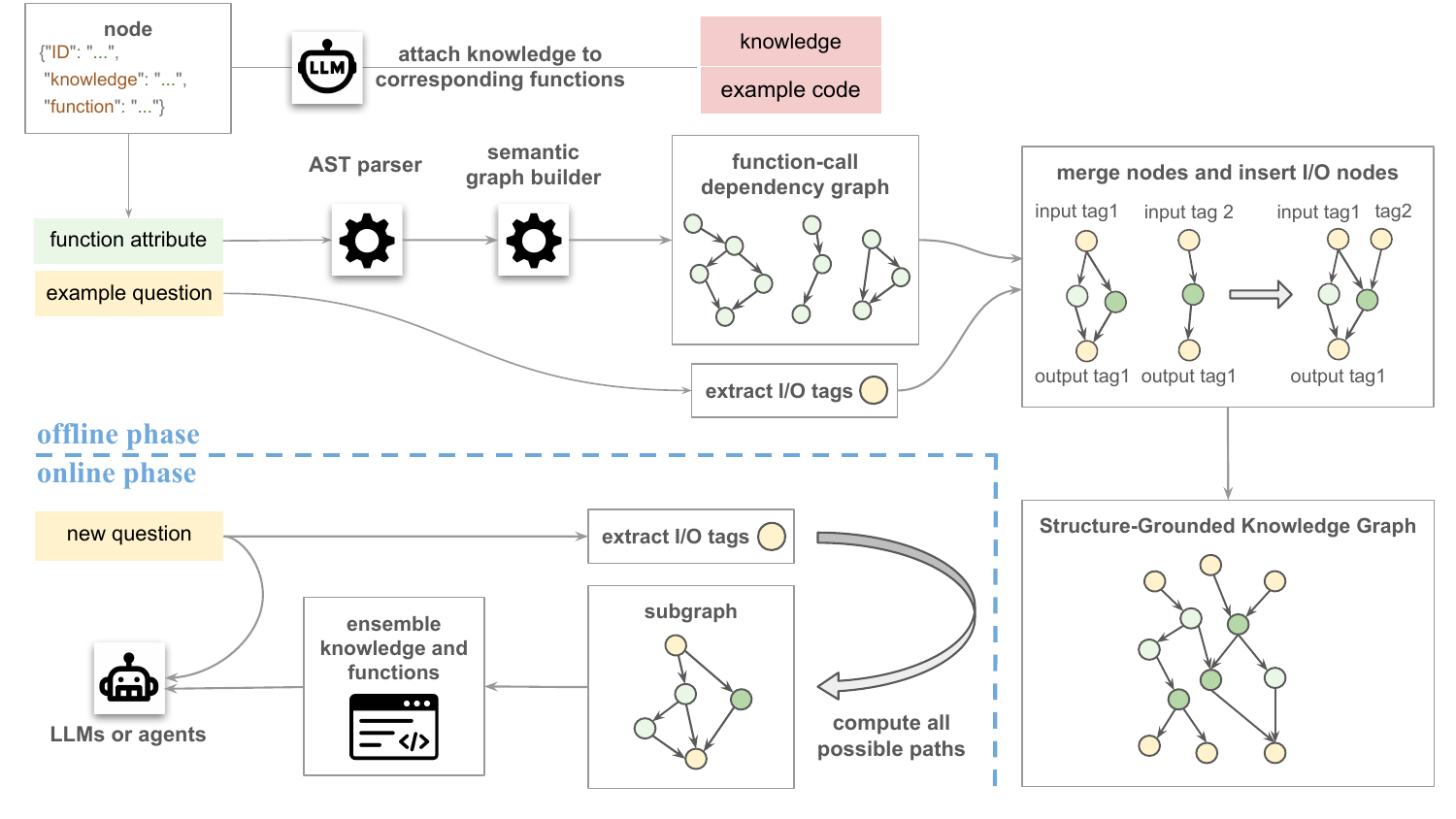}
  \caption{Framework of \textsc{SGKR}. Framework of the structure-grounded code retrieval system. In the offline phase, code is parsed via AST to build a semantic graph and a function-call dependency graph, while external knowledge is inserted into relevant functions as comments. Extracted sementic input–output (I/O) tags are added as nodes to augment the graph. At inference time, I/O tags from a query guide the retrieval of relevant subgraphs through dependency paths, and the retrieved functions and knowledge are provided to LLMs for reasoning.}
  \label{fig:motivation}
\end{figure*}

\section{Structure-Grounded Knowledge Graph}

In this section, we show how to ground abstract knowledge as a directed acyclic graph (DAG) through function dependencies, where each piece of knowledge is mapped to its corresponding function.

\subsection{RQ1: How to define a knowledge unit?}

Standard RAG treats knowledge as unstructured text and relies on embedding-based retrieval. However, in data analysis tasks, knowledge is often operationalized through executable functions that perform intermediate computations. Therefore, representing knowledge purely as text fails to capture its functional role in reasoning.

To better align with programmatic reasoning, we define a \textit{knowledge unit} as a function together with its associated domain knowledge. By binding knowledge to functions, each function serves as a carrier of domain knowledge, while function calls encode how these knowledge units interact. This representation allows knowledge to be organized according to executable structure rather than isolated textual fragments.

\subsection{RQ2: How is knowledge connected?}

After binding functions with their associated knowledge, functions become carriers of domain knowledge, and function calls define explicit connections between knowledge units. Although knowledge itself cannot always be precisely parsed, the function code that implements it exposes clear dependency structures. As a result, relevant knowledge is implicitly connected through code dependencies rather than surface-level similarity. 

If a function $f_1$ calls $f_2$, we add a directed edge $f_1 \rightarrow f_2$, indicating that $f_1$ depends on the knowledge implemented in $f_2$. This structure naturally captures how knowledge is propagated across computation steps.

\subsection{RQ3: How to measure relevance of knowledge?}

Standard similarity-based retrieval assumes that relevant knowledge is semantically similar to the query. However, this assumption often fails in programmatic data analysis. As illustrated in Figure~\ref{fig:motivation} using an example from the \textsc{DABstep}~\cite{dabstep} benchmark, functions with high embedding similarity (e.g., \texttt{check\_month\_volumn}) may be irrelevant to the task, while necessary functions (e.g., \texttt{check\_intracountry}), which encode critical domain knowledge (e.g., \textit{fees depend on whether a transaction is international}), may have low similarity scores and be overlooked.


In contrast, we observe that the needed knowledge is correctly captured by the common function dependency between $Q_1$ and $Q_2$, as shown in dependency graph of Figure~\ref{fig:motivation}. The directed graph illustrates the function-call structure of the solution to $Q_1$ (not shown in full for conciseness), with $Q_1$'s input/output attributes attached. The bold nodes and arrows highlight the functions required to solve $Q_2$, and the input/output attributes of this question. The path connecting the input/output attributes of $Q_2$ covers the knowledge required to solve it.

\subsection{RQ4: How to Enable Efficient Retrieval?}

We introduce semantic input/output (I/O) tags to explicitly represent the I/O entities of a question, mapping to I/O nodes of paths in the graph. 

Semantic I/O tags represent semantic elements involved in the analysis, including data entities (e.g., merchant), attributes (e.g., account type), and target-derived quantities (e.g., average fee, applicable IDs). 

Semantic I/O nodes are inserted into the dependency graph as explicit input and output entities, serving as semantic anchors that guide retrieval and define valid reasoning paths over the graph.





\section{Approach}

We propose \textbf{Structure-Grounded Knowledge Retrieval} (\textsc{SGKR}), a retrieval framework for domain-specific data analysis. The overall pipeline consists of two stages: (1) \textit{offline graph construction}, where we build a structure-grounded graph with code and associated knowledge, and (2) \textit{online retrieval}, where we perform a breadth-first search (BFS) on the graph to identify dependency paths between semantic input and output nodes and retrieve the knowledge-code nodes along those paths.

\subsection{Problem Formulation}


Given a natural language question $q$, the goal is to retrieve a subset of nodes $S \subseteq V_{KC}$ whose associated code and knowledge provide the most useful context for 
a downstream language model to solve the task.

Conceptually, the retrieval target is to select a node set $S$ that maximizes the task performance of the model:

\begin{equation}
S^* = \arg\max_{S \subseteq V_{KC}} \text{Acc}(q, S),
\end{equation}

\noindent where $\text{Acc}(q, S)$ denotes the task performance achieved when answering question $q$ using the retrieved knowledge-code nodes in $S$ as context.

\subsection{Structure-Grounded Knowledge Graph Construction}
\label{sec:graph_construction}

In the offline stage, we construct a graph to organize domain knowledge by extracting function-call relationships from example code snippets.

\paragraph{Data Preparation and Knowledge Attachment}

To ground domain knowledge into the code dependency graph, we represent each node using three components: a node identifier, the function code, and the associated knowledge. 
When a function does not require explicit domain knowledge, we instead attach a brief description of its functionality. In this way, each node encodes both executable logic and the semantic information needed to interpret its role in task solving.

\paragraph{Dependency Extraction via Function Calls}
\label{sec:semantic_parser}

To extract dependency relations between functions, we first parse the source code into an Abstract Syntax Tree (AST) and collect all function definitions. For each function, we traverse its AST to identify function call expressions and establish directed edges from the calling function to the called function, thereby capturing all function dependencies in the code.


\paragraph{Identical Nodes Merging and Semantic I/O Node Insertion}
\label{sec:merging}


The parser described in Section~\ref{sec:semantic_parser} may 
produce multiple nodes that correspond to the same function, since the same 
function can appear in different isolated execution traces. To obtain a consistent graph structure, we first merge such duplicate nodes. Specifically, nodes with the 
same function name are replaced with a single canonical node, and all incoming 
and outgoing edges of the duplicate nodes are redirected to this node. This 
ensures that each function is represented uniquely in the graph while 
preserving all dependency relations.

After merging identical function nodes, we insert the extracted semantic I/O nodes and connect 
them to the corresponding function nodes according to the parsed dependencies. 
This allows the graph to capture potential relations between different
inputs and outputs that share the same functions. For example, suppose the parsed 
dependencies contain the paths
\textit{input$_1$ $\rightarrow$ func$_1$ $\rightarrow$ output$_1$} and 
\textit{input$_2$ $\rightarrow$ func$_1$ $\rightarrow$ output$_2$}. After merging 
duplicate instances of \textit{func$_1$}, the graph can additionally expose the potential 
structural connection \textit{input$_1$ $\rightarrow$ func$_1$ $\rightarrow$ output$_2$}.

The overall procedure is summarized in Algorithm~\ref{alg:io_insertion}. We further analyze the effect of this merging strategy in the ablation study in Section~\ref{sec:ablation}.

\begin{algorithm}[t]
\caption{Identical Node Merging and I/O Node Insertion}
\label{alg:io_insertion}
\begin{algorithmic}[1]
\REQUIRE Dependency graph $G=(V,E)$, I/O nodes $V_{I/O}$
\ENSURE Updated graph $G'=(V',E')$ with unique function nodes and $V_{I/O}$ inserted

\STATE $G' \leftarrow G$

\STATE Group identical nodes in $V'$

\FOR{each group $S = \{v_1, v_2, ..., v_k\}$ representing the same function}
    \STATE Select one node $v_c$ as the canonical node
    \FOR{each duplicate node $v_i \in S \setminus \{v_c\}$}
        \STATE Redirect incoming edges: $(u, v_i) \rightarrow (u, v_c)$
        \STATE Redirect outgoing edges: $(v_i, w) \rightarrow (v_c, w)$
        \STATE Remove $v_i$ from $V'$
    \ENDFOR
\ENDFOR

\FOR{each I/O node $v \in V_{I/O}$}
    \STATE Add $v$ to $V'$
    \IF{$v$ is an input node}
        \STATE Find the first function node $f$ that consumes $v$
        \STATE Add edge $(v, f)$
    \ELSIF{$v$ is an output node}
        \STATE Find the function node $f$ that returns $v$
        \STATE Add edge $(f, v)$
    \ENDIF
\ENDFOR

\RETURN $G'=(V',E')$
\end{algorithmic}
\end{algorithm}

\subsection{Retrieval via BFS graph search}

Given a new question, we perform graph-based retrieval to identify relevant knowledge and corresponding functions.

\paragraph{Semantic I/O Tag Extraction}

As discussed in Section~\ref{sec:merging}, retrieval begins by matching the question to semantic I/O nodes in the graph. For each new question, we extract semantic input/output (I/O) tags using keyword matching, and use these tags to identify the corresponding semantic I/O nodes in the knowledge--code dependency graph. 
If no tags can be identified, the system falls back to the vanilla inference pipeline without retrieval. 


\paragraph{Subgraph Retrieval}
\label{sec:subgraph}

Given the identified semantic I/O nodes, we retrieve a task-relevant subgraph from the full structure-grounded knowledge--code graph. 
Starting from the input nodes, we perform breadth-first search (BFS) to identify dependency paths that connect them to the output nodes. During traversal, neighboring nodes are expanded while the current search paths are tracked. Whenever the search reaches an output node, the corresponding path is recorded as a dependency chain linking the input concepts to the desired outputs. The union of nodes appearing on the discovered paths forms the retrieved subgraph. 
Empirically, the BFS returns only a small number of dependency paths for each query, which keeps the retrieved subgraph compact.


\paragraph{Context Construction from Retrieved Subgraph}
Finally, we traverse the retrieved subgraph obtained from steps in Section~\ref{sec:subgraph} and assemble the context provided to the language model. We ignore semantic I/O nodes and extract the associated information from the remaining knowledge-code nodes. Specifically, the \textit{knowledge} fields are concatenated to form the textual context, while the \textit{function} fields are collected as executable code examples. The resulting context combines domain knowledge with executable function implementations derived from the retrieved subgraph, which is supplied to the language model to support downstream reasoning.


\begin{table*}[t]
\centering
\resizebox{0.92\textwidth}{!}{
\begin{tabular}{l cc cc cc}
\hline
\multirow{2}{*}{\textbf{Method}} & 
\multicolumn{2}{c}{\textbf{DABstep-hard}} & 
\multicolumn{2}{c}{\textbf{ConvFinQA-hard}} & 
\multicolumn{2}{c}{\textbf{FinQA-hard}} \\
 & \textbf{Acc.} & \textbf{Avg. Nodes} &
   \textbf{Acc.} & \textbf{Avg. Nodes} &
   \textbf{Acc.} & \textbf{Avg. Nodes} \\
\hline
vanilla & 6.88\% & -- & 52.00\% & -- & 47.50\% & -- \\
few-shot & 11.64\% & -- & 63.00\% & -- & 57.78\% & -- \\
\textsc{bge-large-en-v1.5-RAG} & 10.58\% & 5 & 71.33\% & 2 & 63.61\% & 2 \\
\textsc{GraphSAGE-RAG} & 6.35\% & 5 & 71.67\% & 2 & 62.78\% & 2 \\
\textsc{CodeBERT-RAG} & 8.73\% & 5 & 68.33\% & 2 & 61.94\% & 2 \\
\textsc{SGKR} & \textbf{24.34\%} & \textbf{4.65} & \textbf{74.67\%} & \textbf{1.52} & \textbf{65.56\%} & \textbf{1.50} \\
\hline
\end{tabular}
}
\caption{Accuracy comparison across datasets and average number of retrieved nodes (semantic I/O nodes are not counted).}
\label{tab:main_res}
\end{table*}

\begin{table}[t]
\centering
\small
\begin{tabular}{|l|c|}
\hline
\textbf{Method} & \textbf{Acc.} \\
\hline
No Retrieval & 6.88\% \\
\textsc{SGKR} w/ Fixed Nodes & 11.64\% \\
\textsc{SGKR} w/o Knowledge & 8.73\% \\
\textsc{SGKR} w/o Merging & 12.43\% \\
\hline
Full \textsc{SGKR} & 24.34\% \\
\hline
\end{tabular}
\caption{Ablation study of \textsc{SGKR}. Each variant removes one key component of the proposed framework.}
\label{tab:ablation}
\end{table}

\begin{table*}[h]
\centering
\resizebox{\textwidth}{!}{
\begin{tabular}{lccccc}
\hline
\textbf{knowledge \& function} & \textbf{Ground Truth} & \textsc{SGKR} & \textsc{GraphSAGE-RAG} & \textsc{CodeBERT-RAG} & \textsc{dense-RAG} \\
\hline
\texttt{rule\_applies}            & needed & \checkmark & \checkmark & \checkmark &  \\
\texttt{find\_all\_mccs}          & needed & \checkmark &            &            &  \\
\texttt{sum\_fee}                 & needed & \checkmark &            &            &  \\
\texttt{most\_expensive}          & needed & \checkmark & \checkmark &            & \checkmark \\
\texttt{compute\_fee}             & needed & \checkmark & \checkmark &            &  \\
\texttt{average\_fee}             & unneeded           &            & \checkmark & \checkmark & \checkmark \\
\texttt{output\_average\_fee}     &  unneeded          &            & \checkmark &            &  \\
\texttt{merchant\_matches\_fee}   &  unneeded          &            &            & \checkmark &  \\
\texttt{match\_fee\_conditions} &    unneeded        &            &            & \checkmark & \checkmark \\
\texttt{match\_capture\_delay}  &  unneeded          &            &            & \checkmark &  \\
\texttt{cheapest\_card\_scheme}   &  unneeded          &            &            &            & \checkmark \\
\texttt{get\_mcc\_code\_from\_dsp}& unneeded           &            &            &            & \checkmark \\
\hline
\end{tabular}
}
\caption{Retrieved knowledge-code nodes by different methods. \checkmark means the knowledge is selected by the method.}
\label{tab:case_study}
\end{table*}

\section{Experiment}

\subsection{Datasets}
We evaluate \textsc{SGKR} on two domain-specific benchmarks that require multi-step reasoning and code-based problem solving.

\begin{enumerate}
    \item \textsc{DABstep}~\cite{dabstep} is a benchmark for realistic multi-step data analysis tasks. It contains over 450 real-world challenges that require models to perform code-based data processing and contextual reasoning over heterogeneous documentation.
    \item \textsc{ConvFinQA}~\cite{ConvFinQA} and FinQA~\cite{finqa} are benchmarks for multi-step numerical reasoning in the financial domain. \textsc{ConvFinQA} focuses on conversational question answering, while \textsc{FinQA} consists of standalone questions. Both datasets require models to perform multi-step numerical reasoning over financial tables and associated textual context.
\end{enumerate}

We focus on domain-specific datasets because \textsc{SGKR} retrieves knowledge from a dependency graph constructed from domain code and its associated semantics. For such retrieval to be meaningful, benchmark questions must require reasoning grounded in the same domain knowledge encoded in the graph. In contrast, general programming tasks or open-ended Kaggle-style data science problems are typically less centered around a shared body of domain code and operational knowledge, making them less suitable for evaluating our setting.

Since our goal is to evaluate multi-step reasoning over code dependencies, we restrict evaluation to \textbf{hard} questions in each dataset, defined as instances whose gold solutions require more than one reasoning/computation step. After filtering, \textsc{DABstep} remains 378 questions, \textsc{FinQA} remains 360 questions, and \textsc{ConvFinQA} remains 300 questions.

\subsection{Baselines}

We compare \textsc{SGKR} with several representative baselines:

\begin{enumerate}
\item \textbf{Vanilla Prompting}: The model answers the question without retrieved context.

\item \textbf{Few-shot Prompting}: The model is provided with a fixed set of demonstration examples in the prompt.

\item \textsc{Dense-RAG}: Retrieves knowledge--code nodes using \texttt{bge-large-en-v1.5}~\cite{bge_embedding}, a strong general-purpose dense retriever widely adopted in RAG systems.

\item \textsc{CodeBERT-RAG}: Retrieves nodes using \textsc{CodeBERT}~\cite{codebert}, a code-pretrained model designed for code understanding and retrieval.

\item \textsc{GraphSAGE-RAG}~\cite{gragc}: 
Learns node representations with \textsc{GraphSAGE} and retrieves nodes according to their similarity to the query representation.
\end{enumerate}


Unlike graph retrieval settings that assume a predefined knowledge graph, our setting requires constructing a knowledge--code dependency graph directly from executable code. To enable a fair comparison, all retrieval methods operate on the same graph. This design isolates the effect of the retrieval strategy itself, rather than mixing it with variances in knowledge-source construction or graph building. We note that knowledge-source construction is an important design choice in its own right; our goal here is to compare retrieval strategies using under a shared graph structure.


\subsection{Experimental Preparation and Settings}


Since \textsc{SGKR} retrieves a variable number of nodes depending on the discovered subgraph, we set the baseline retrieval budget to match its average retrieved context size on each dataset, enabling comparison under similar context budgets. This yields a top-$5$ setting on DABStep and a top-$2$ setting on ConvFinQA and FinQA.



For \textsc{FinQA} and \textsc{ConvFinQA}, we convert the numerical computation symbols in the original annotations into Python functions. Since these datasets provide annotated computation programs together with relevant financial domain knowledge, we directly attach the corresponding domain knowledge to each computation function when constructing the graph.

For \textsc{DABStep}, only a limited number of examples are publicly released, and they do not include complete solution processes. We therefore use an LLM (\texttt{GPT-5.2}) to generate step-by-step code solutions, while incorporating relevant domain knowledge from the benchmark's provided documentation into the corresponding functions. These generated solutions are then validated through code execution to ensure correctness before being used for graph construction.

For DABstep, we adopt the ReAct agent workflow built upon \texttt{smolagents}, as implemented in the official benchmark repository.\footnote{\url{https://huggingface.co/spaces/adyen/DABstep/tree/main/baseline}}

\subsection{Main Results}

Table~\ref{tab:main_res} reports the performance of different retrieval strategies on the hard subsets of \textsc{DABstep}, \textsc{ConvFinQA}, and \textsc{FinQA}. We also report the average number of retrieved nodes. 
Overall, \textsc{SGKR} consistently achieves the best performance across all three datasets. On \textsc{DABstep}-hard, \textsc{SGKR} improves accuracy to 24.34\%, substantially outperforming all baselines. This result highlights both the difficulty of the benchmark and the value of structure-grounded retrieval in complex multi-step data analysis tasks.



On ConvFinQA-hard and FinQA-hard, \textsc{SGKR} also consistently outperforms all baselines, while retrieving approximately 25\% fewer nodes on average than other embedding-based retrieval methods.

Taken together, these results support our intuition that code dependency structure provides a stronger retrieval signal than surface-level similarity for domain-specific multi-step reasoning tasks. As a result, the proposed Structure-Grounded retrieval mechanism more effectively identifies task-relevant knowledge, leading to higher accuracy with a more compact retrieved context.

\subsection{Ablation Study}
\label{sec:ablation}

To understand the contribution of each component in \textsc{SGKR}, we conduct an ablation study by progressively removing key modules from the full system.

\paragraph{No Retrieval.}
In this setting, the model solves the task without accessing the code graph or retrieving any external knowledge. As shown in Table~\ref{tab:ablation}, performance drops significantly to 6.88\%, indicating that external structured knowledge is crucial for solving complex reasoning tasks.

\paragraph{\textsc{SGKR} with Fixed Nodes.}
Instead of dynamically constructing a query-specific subgraph and retrieving relevant knowledge and functions from graph nodes, we use a static set of frequently used nodes. This configuration achieves 11.64\% accuracy, substantially lower than the full system, highlighting the importance of question-specific subgraph retrieval.

\paragraph{\textsc{SGKR} without Knowledge.}
To isolate the contribution of structured knowledge beyond function code alone, we remove the knowledge attached to retrieved nodes while retaining the corresponding functions in the retrieved subgraph. The resulting performance degradation indicates that structured knowledge, in addition to code-level dependency information, is critical to \textsc{SGKR}.

\paragraph{\textsc{SGKR} without Merging Identical Nodes.}
As discussed in Section~\ref{sec:merging}, the AST parser may introduce duplicate nodes when the same function appears in different isolated execution traces. Without merging such nodes, \textsc{SGKR} may retrieve redundant copies of the same function and its associated knowledge. Likewise, similarity-based retrieval methods may repeatedly select duplicate nodes with high similarity scores, leading to redundant and less informative retrieval results.

\paragraph{Full \textsc{SGKR}}
The full system integrates structure-grounded graph retrieval, dynamic semantic I/O node insertion, and executable code demonstrations. This configuration achieves the best performance, reaching 24.34\% accuracy. The result confirms that structure-grounded retrieval over code dependency structure is important for effective multi-step reasoning.

\subsection{Case Study}

We present a case study on the question: \textit{``What is the most expensive MCC for a transaction of 5 euros, in general?''} from \textsc{DABStep}, for which only only \textsc{SGKR} produces the correct solution. Table~\ref{tab:case_study} shows the code snippets retrieved by different methods against the ground truth computation pipeline.

Baseline retrieval methods are often misled by lexical similarity in function names or attributes. For example, \texttt{merchant\_matches\_fee()} appears highly relevant because the question mentions both \emph{merchant category codes (MCC)} and \emph{fee}. However, this function is actually designed to filter fees for specific merchants and depends on additional attributes beyond \emph{MCC}, making it irrelevant to the required reasoning chain. Similarly, \texttt{cheapest\_card\_scheme()} is retrieved due to the semantic similarity between \emph{cheapest} and \emph{most expensive}, even though \emph{card scheme} is not used in the fee-rule filtering logic needed for this task.

In contrast, \textsc{SGKR} retrieves the functions that form the actual computation pipeline, including rule matching, MCC enumeration, and fee computation. This example illustrates that retrieval based purely on semantic similarity can be distracted by superficially related identifiers, whereas structure-grounded retrieval can better recover the task-critical domain knowledge required for correct reasoning.






\section{Related Work}

\textbf{LLMs for Code Generation in Data Analysis.} Large language models (LLMs) have shown strong capability in generating executable code for solving complex reasoning and data analysis tasks~\cite{tang2025analyst,perez2025llminsight,hong2025data}. These tasks often require combining domain knowledge with correct sequences of computational operations. Recent work has demonstrated that LLMs can translate natural language queries into executable programs for tasks such as data analysis, data management and database querying~\cite{dsbench,jiang2025sqlgovernorllmpoweredsqltoolkit}. However, LLMs often struggle to select the knowledge required to solve a new problem, especially when the relevant knowledge is not lexically similar to the new question~\cite{arming}.

\textbf{Graph-Based Retrieval.} Recent work explores graph-based retrieval to better capture structured relationships in knowledge~\cite{liang2026sentgraphhierarchicalsentencegraph,saleh2024sg}. Instead of retrieving documents based solely on embedding similarity, these approaches organize knowledge into graphs and perform multi-hop retrieval over the graph structure. Such methods have been applied to knowledge-intensive tasks such as multi-hop question answering~\cite{graphrag,yang2018hotpotqa,shavaki2024knowledgerag}, where reasoning requires connecting multiple pieces of information. In contrast, our work focuses on the generation of data analysis code, where the key challenge is identifying the relevant knowledge and the sequence of computational operations required to transform inputs into outputs.

\textbf{Graph-Based Program Representation} Prior work represents programs as graphs, such as abstract syntax trees~\cite{allamanis2018learning}, data-flow graphs~\cite{graphcodebert,cummins2021programl}, and code property graphs, to capture structural relationships between program components. These representations are primarily developed for program analysis and code searching tasks~\cite{deep}. In contrast, our work leverages program structure as a carrier of knowledge, constructing a dependency graph that supports knowledge retrieval for data analysis code generation.





\section{Conclusion}
\label{sec:bibtex}

We propose Structure-Grounded Knowledge Retrieval (\textsc{SGKR}), a structured-grounded framework to retrieve knowledge from previous code snippets. Instead of relying on embedding similarity, \textsc{SGKR} leverages function-call relationships to identify knowledge that forms executable reasoning paths. By representing domain knowledge as a code dependency graph and retrieving dependency-consistent subgraphs, \textsc{SGKR} provides structured grounding for LLM-based code generation. Experiments on multiple multi-step data analysis benchmarks demonstrate that \textsc{SGKR} consistently improves solution correctness compared to embedding-based retrieval for both vanilla LLMs and coding agents.

\section{Acknowledgments}

The author gratefully acknowledges Mingzhe Lu, Haoyu Dong and Zhengjie Miao for constructive feedback on this work.

\clearpage

\section*{Limitations and Future Directions}

Despite its effectiveness, \textsc{SGKR} has several limitations.

First, our approach assumes the availability of executable example code and associated domain knowledge that can be organized into a dependency graph. Constructing such resources may require additional effort when adapting the system to new domains. Moreover, \textsc{SGKR} is most effective for tasks that rely on domain knowledge and reusable computational patterns, and may provide limited benefits for code-generation tasks that do not require external knowledge. 

Second, the current implementation relies on a semantic parser that extracts dependencies from compilable code. For datasets such as FinQA and ConvFinQA, where solutions consist of symbolic numerical expressions rather than executable code, we must convert these expressions into Python function calls. This additional wrapping step may introduce engineering overhead and could limit direct applicability to tasks where reasoning steps are not naturally expressed as executable programs.

For future work, we plan to explore automatic generation of such example programs and associated knowledge using large language models. For instance, a pre-play or simulation stage could generate candidate reasoning programs before inference, which can then be incorporated into the dependency graph to expand coverage. This direction could reduce the need for manually curated examples and improve the scalability of the proposed framework across domains.

\section*{Ethical Considerations}

\paragraph{Risk of over-trusting LLM outputs.}
While \textsc{SKGR} provides structured grounding through code dependencies, the generated reasoning and code may still be incorrect. The explicit dependency structure and executable form can create a false sense of reliability, leading users to over-trust the outputs. This risk is particularly important in high-stakes domains such as healthcare. Users should treat generated results as assistive rather than authoritative and incorporate validation or human oversight when necessary.


\paragraph{Use of AI assistants.}
AI assistants were used in the development of this work to support code generation, writing, and editing. All outputs were reviewed, validated, and refined by the authors to ensure correctness and clarity. The final content reflects the authors' own understanding and contributions.

\bibliography{custom}

\appendix

\section{Appendix}
\subsection{Prompts for \textsc{DABstep}}
\label{sec:prompts}

\begin{tcolorbox}[title=Prompt of \textsc{DABStep} Evaluation, colback=gray!5, colframe=gray!60, boxrule=0.5pt, breakable]

\textbf{System Prompt}

You are an expert data analyst who can solve any task using code blobs. You will be given a task to solve as best as you can.
In the environment there exists data which will help you solve your data analyst task. This data is spread out across files in the directory \texttt{\{ctx\_path\}}.

For each task you will follow a hierarchy of workflows: a root-level task workflow and a leaf-level step workflow.
There is one root-level workflow per task which consists of multiple leaf-level step workflows.

\textbf{Root Task Workflow:} 

\textit{Explore} $\rightarrow$ \textit{Plan} $\rightarrow$ \textit{Execute} $\rightarrow$ \textit{Conclude}

\begin{enumerate}
\item \textbf{Explore:} Perform data exploration in the directory \texttt{\{ctx\_path\}} and understand what data is available and its limitations.
\item \textbf{Plan:} Draft a high-level plan based on the results of the Explore step.
\item \textbf{Execute:} Execute the drafted plan. If the plan fails, restart from the Explore step.
\item \textbf{Conclude:} Based on the executed plan, summarize the findings into an answer for the task.
\end{enumerate}

\textbf{Step Workflow:} \textit{Thought} $\rightarrow$ \textit{Code} $\rightarrow$ \textit{Observation}

\begin{enumerate}
\item \textbf{Thought:} Explain your reasoning and the code you will use.
\item \textbf{Code:} Write Python code in the following format:

\begin{verbatim}
Code:
```py
your_python_code

<end_code>
\end{verbatim}

Use \texttt{print()} to retain important outputs.

\item \textbf{Observation:} Review the printed outputs before continuing.
\end{enumerate}

\textbf{Rules}

\begin{itemize}
\item Always check the directory \texttt{ctx\_path} for relevant documentation or data before assuming information is unavailable.
\item Validate assumptions using the available documentation before executing code.
\item If all possible solution plans fail, output ``Not Applicable'' as the final answer.
\item Use only defined variables and valid Python statements.
\item Avoid long unpredictable code snippets in a single step.
\item Use Python only when necessary and avoid repeating code that previously failed.
\item Do not create notional variables.
\item Imports and variables persist between executions.
\item Solve the task directly instead of giving instructions.
\item Allowed imports: \texttt{authorized\_imports}.
\item Never import \texttt{final\_answer}; it is already available.
\end{itemize}

\textbf{Available Tool}

\texttt{\{final\_answer()\}}

Example:

\begin{verbatim}
answer = df["result"].mean()
final\_answer(answer)
<end\_code>
\end{verbatim}

\textbf{Task Instruction}

\texttt{\{question\}}

You must follow these guidelines when producing your final answer:

\texttt{\{guidelines\}}

\textbf{Context and Hints}

Following domain knowledge and context should be helpful:

\texttt{\{retrieved\_knowledge\}}

Here are some example functions that you may refer to:

\texttt{\{retrieved\_functions\}}

\end{tcolorbox}

\subsection{Prompts for ConvFinQA and FinQA}

\begin{tcolorbox}[
title=Prompt for \textsc{ConvFinQA} and \textsc{FinQA} Evaluation,
colback=gray!5, colframe=gray!60, boxrule=0.5pt, breakable
]

You are an expert in financial reasoning.

\textbf{Output Requirement}

Return \textbf{ONLY} the program tokens as a JSON list, ending with \texttt{"EOF"}.

Format:

\begin{verbatim}
["operation(", "val1", "val2", ")", 
"EOF"]
\end{verbatim}

\textbf{Multi-step Calculation Rules}

\begin{itemize}
\item Write operations sequentially (not nested).
\item Use \#0, \#1, \#2, etc. to reference previous results.
\item \#0 = result of first operation, \#1 = result of second operation.
\end{itemize}

\textbf{Token Format Rules}

\begin{itemize}
\item Each token must be a separate string in the JSON list.
\item Numbers must appear as plain numbers (e.g., \texttt{"123"}).
\item Operations must end with parentheses: \texttt{"add("}, \texttt{"subtract("}, \texttt{"multiply("}, \texttt{"divide("}, \texttt{"greater("}.
\item Closing parentheses must be separate tokens: \texttt{")"}.
\item Always end with \texttt{"EOF"}.
\item For percentages, do not convert decimals to percent format (e.g., use \texttt{"0.123"} for 12.3\%).
\item For constants, use predefined tokens (e.g., \texttt{const\_100}).
\end{itemize}

\textbf{Supported Operations}

Basic math

\begin{itemize}
\item add(val1, val2)
\item subtract(val1, val2)
\item multiply(val1, val2)
\item divide(val1, val2)
\end{itemize}

Comparison

\begin{itemize}
\item greater(val1, val2) — returns 1 if val1 $>$ val2, else 0
\end{itemize}

\textit{Constants}

\begin{itemize}
\item const\_2, const\_10, const\_100000
\item const\_m1
\end{itemize}

\textbf{Examples}

\begin{verbatim}
Single-step:
["divide(", "8.1", "56.0", ")", "EOF"]

Multi-step:
["subtract(", "153.7", "139.9", ")", 
 "divide(", "#0", "139.9", ")", 
 "EOF"]

With constants:
["multiply(", "#1", "const_100", ")", 
"EOF"]

\end{verbatim}

\textbf{Context and Hints}

Following domain knowledge and context should be helpful:

\texttt{\{retrieved\_knowledge\}}

Here are some example functions that you may refer to:

\texttt{\{retrieved\_functions\}}

\end{tcolorbox}



\end{document}